\documentclass[runningheads,a4paper]{llncs}

\usepackage{amsmath}
\usepackage{amssymb}
\usepackage[misc]{ifsym}
\usepackage{graphicx}
\usepackage{color}

\usepackage{algorithm}
\usepackage{algorithmicx}
\usepackage{algpseudocode}
\usepackage{multirow}
\usepackage{array}
\newcolumntype{L}[1]{>{\raggedright\let\newline\\\arraybackslash\hspace{0pt}}m{#1}}
\newcolumntype{C}[1]{>{\centering\let\newline\\\arraybackslash\hspace{0pt}}m{#1}}
\newcolumntype{R}[1]{>{\raggedleft\let\newline\\\arraybackslash\hspace{0pt}}m{#1}}

\usepackage{url}
\urldef{\mailsa}\path|{alfred.hofmann, ursula.barth, ingrid.haas, frank.holzwarth,|
\urldef{\mailsb}\path|anna.kramer, leonie.kunz, christine.reiss, nicole.sator,|
\urldef{\mailsc}\path|erika.siebert-cole, peter.strasser, lncs}@springer.com|

\begin{document}

\mainmatter  

\title{Bridging the Gap Between 2D and 3D \\ Organ Segmentation with Volumetric Fusion Net}

\titlerunning{Bridging the Gap Between 2D and 3D Organ Segmentation}

%
%
\author{Yingda Xia\inst{1}, Lingxi Xie\inst{1}$^{(\textrm{\Letter})}$, Fengze Liu\inst{1}, Zhuotun Zhu\inst{1}, \\Elliot K. Fishman\inst{2}, Alan L. Yuille\inst{1}}
\authorrunning{Y. Xia \textit{et al.}}

\institute{The Johns Hopkins University, Baltimore, MD 21218, USA \\ \and
The Johns Hopkins University School of Medicine, Baltimore, MD 21287, USA
}

%
%

\toctitle{Bridging the Gap Between 2D and 3D Organ Segmentation}
\tocauthor{Anonymous Authors}
\maketitle

\begin{abstract}
There has been a debate on whether to use 2D or 3D deep neural networks for volumetric organ segmentation. Both 2D and 3D models have their advantages and disadvantages. In this paper, we present an alternative framework, which trains 2D networks on different viewpoints for segmentation, and builds a 3D {\bf Volumetric Fusion Net} (VFN) to fuse the 2D segmentation results. VFN is relatively shallow and contains much fewer parameters than most 3D networks, making our framework more efficient at integrating 3D information for segmentation. We train and test the segmentation and fusion modules individually, and propose a novel strategy, named {\em cross-cross-augmentation}, to make full use of the limited training data. We evaluate our framework on several challenging abdominal organs, and verify its superiority in segmentation accuracy and stability over existing 2D and 3D approaches.
\end{abstract}

\vspace{-0.4cm}
\section{Introduction}
\label{Introduction}
\vspace{-0.2cm}

With the increasing requirement of fine-scaled medical care, computer-assisted diagnosis (CAD) has attracted more and more attention in the past decade. An important prerequisite of CAD is an intelligent system to process and analyze medical data, such as CT and MRI scans. In the area of medical imaging analysis, organ segmentation is a traditional and fundamental topic~\cite{boykov2000interactive}. Researchers often designed a specific system for each organ to capture its properties. In comparison to large organs ({\em e.g.}, the liver, the kidneys, the stomach, {\em etc.}), small organs such as the pancreas are more difficult to segment, which is partly caused by their highly variable geometric properties~\cite{roth2015deeporgan}.

In recent years, with the arrival of the deep learning era~\cite{krizhevsky2012imagenet}, powerful models such as convolutional neural networks~\cite{long2015fully} have been transferred from natural image segmentation to organ segmentation. But there is a difference. Organ segmentation requires dealing with volumetric data, and two types of solutions have been proposed. The first one trains 2D networks from three orthogonal planes and fusing the segmentation results~\cite{roth2015deeporgan}\cite{yu2017recurrent}\cite{zhou2017fixed}, and the second one suggests training a 3D network directly~\cite{cicek20163d}\cite{milletari2016v}\cite{zhu20173d}. But 3D networks are more computationally expensive yet less stable when trained from scratch, and it is difficult to find a pre-trained model for medical purposes. In the scenario of limited training data, fine-tuning a pre-trained 2D network~\cite{long2015fully} is a safer choice~\cite{tajbakhsh2016convolutional}.

This paper presents an alternative framework, which trains 2D segmentation models and uses a light-weighted 3D network, named {\bf Volumetric Fusion Net} (VFN), in order to fuse 2D segmentation at a late stage. A similar idea is studied before based on either the EM algorithm~\cite{asman2013non} or pre-defined operations in a 2D scenario~\cite{yang2016deep}, but we propose instead to construct generalized linear operations (convolution) and allow them to be learned from training data. Because it is built on top of reasonable 2D segmentation results, VFN is relatively shallow and does not use fully-connected layers (which contribute a large fraction of network parameters) to improve its discriminative ability. In the training process, we first optimize 2D segmentation networks on different viewpoints individually (this strategy was studied in~\cite{su2015multi}\cite{setio2016pulmonary}\cite{zhou2017fixed}), and then use the validation set to train VFN. When the amount of training data is limited, we suggest a {\em cross-cross-augmentation} strategy to enable reusing the data to train both 2D segmentation and 3D fusion networks.

We first apply our system to a public dataset for pancreas segmentation~\cite{roth2015deeporgan}. Based on the state-of-the-art 2D segmentation approaches~\cite{yu2017recurrent}\cite{zhou2017fixed}, VFN produces a consistent accuracy gain and outperforms other fusion methods, including majority voting and statistical fusion~\cite{asman2013non}. In comparison to 3D networks such as~\cite{zhu20173d}, our framework achieves comparable segmentation accuracy using fewer computational resources, {\em e.g.}, using $10\%$ parameters and being $3\times$ faster at the testing stage (it only adds $10\%$ computation beyond the 2D baselines). We also generalize our framework to other small organs such as the adrenal glands and the duodenum, and verify its favorable performance.

\vspace{-0.4cm}
\section{Our Approach}
\label{Approach}

\vspace{-0.4cm}
\subsection{Framework: Fusing 2D Segmentation into a 3D Volume}
\label{Approach:Framework}
\vspace{-0.2cm}

We denote an input CT volume by $\mathbf{X}$. This is a $W\times H\times L$ volume, where $W$, $H$ and $L$ are the numbers of voxels along the {\em coronal}, {\em sagittal} and {\em axial} directions, respectively. The $i$-th voxel of $\mathbf{X}$, $x_i$, is the intensity (Hounsfield Unit, HU) at the corresponding position, ${i}={\left(1,1,1\right),\ldots,\left(W,H,L\right)}$. The ground-truth segmentation of an organ is denoted by $\mathbf{Y}^\star$, which has the same dimensionality as $\mathbf{X}$. If the $i$-th voxel belongs to the target organ, we set ${y_i^\star}={1}$, otherwise ${y_i^\star}={0}$. The goal of organ segmentation is to design a function $\mathbf{g}\!\left(\cdot\right)$, so that ${\mathbf{Y}}={\mathbf{g}\!\left(\mathbf{X}\right)}$, with all ${y_i}\in{\left\{0,1\right\}}$, is close to $\mathbf{Y}^\star$. We measure the similarity between $\mathbf{Y}$ and $\mathbf{Y}^\star$ by the Dice-S{\o}rensen coefficient (DSC): ${\mathrm{DSC}\!\left(\mathbf{Y},\mathbf{Y}^\star\right)}={\frac{2\times\left|\mathcal{Y}\cap\mathcal{Y}^\star\right|}{\left|\mathcal{Y}\right|+\left|\mathcal{Y}^\star\right|}}$, where ${\mathcal{Y}^\star}={\left\{i\mid y_i^\star=1\right\}}$ and ${\mathcal{Y}}={\left\{i\mid y_i=1\right\}}$ are the sets of foreground voxels.

There are, in general, two ways to design $\mathbf{g}\!\left(\cdot\right)$. The first one trains a 3D model to deal with volumetric data directly~\cite{cicek20163d}\cite{milletari2016v}, and the second one works by cutting the 3D volume into slices, and using 2D networks for segmentation. Both 2D and 3D approaches have their advantages and disadvantages. We appreciate the ability of 3D networks to take volumetric cues into consideration (radiologists also exploit 3D information to make decisions), but, as shown in Section~\ref{Experiments:Ours}, 3D networks are sometimes less stable, arguably because we need to train all weights from scratch, while the 2D networks can be initialized with pre-trained models from the computer vision literature~\cite{long2015fully}. On the other hand, processing volumetric data ({\em e.g.}, 3D convolution) often requires heavier computation in both training and testing ({\em e.g.}, requiring $3\times$ testing time, see Table~\ref{Tab:ComparisonNIH}).

In mathematical terms, let $\mathbf{X}_l^\mathrm{A}$, ${l}={1,2,\ldots,L}$ be a 2D slice (of $W\times H$) along the {\em axial} view, and ${\mathbf{Y}_l^\mathrm{A}}={\mathbf{s}^\mathrm{A}\!\left(\mathbf{X}_l^\mathrm{A}\right)}$ be the segmentation score map for $\mathbf{X}_l^\mathrm{A}$. $\mathbf{s}^\mathrm{A}\!\left(\cdot\right)$ can be a 2D segmentation network such as FCN~\cite{long2015fully}, or a multi-stage system such as a coarse-to-fine framework~\cite{zhou2017fixed}. Stacking all $\mathbf{Y}_l^\mathrm{A}$'s yields a 3D volume ${\mathbf{Y}^\mathrm{A}}={\mathbf{s}^\mathrm{A}\!\left(\mathbf{X}\right)}$. This slicing-and-stacking process can be performed along each axis independently. Due to the large image variation in different views, we train three segmentation models, denoted by $\mathbf{s}^\mathrm{C}\!\left(\cdot\right)$, $\mathbf{s}^\mathrm{S}\!\left(\cdot\right)$ and $\mathbf{s}^\mathrm{A}\!\left(\cdot\right)$, respectively. Finally, a fusion function $\mathbf{f}\!\left[\cdot\right]$ integrates them into the final prediction:
\begin{equation}
\label{Eqn:Segmentation}
{\mathbf{Y}}={\mathbf{f}\!\left[\mathbf{X},\mathbf{Y}^\mathrm{C},\mathbf{Y}^\mathrm{S},\mathbf{Y}^\mathrm{A}\right]}={\mathbf{f}\!\left[\mathbf{X},\mathbf{s}^\mathrm{C}\!\left(\mathbf{X}\right),\mathbf{s}^\mathrm{S}\!\left(\mathbf{X}\right),\mathbf{s}^\mathrm{A}\!\left(\mathbf{X}\right)\right]}.
\end{equation}
Note that we allow the image $\mathbf{X}$ to be incorporated. This is related to the idea known as auto-contexts~\cite{tu2010auto} in computer vision. As we shall see in experiments, adding $\mathbf{X}$ improves the quality of fusion considerably. Our goal is to equip $\mathbf{f}\!\left[\cdot\right]$ with partial abilities of 3D networks, {\em e.g.}, learning simple, local 3D patterns.

\vspace{-0.4cm}
\subsection{Volumetric Fusion Net}
\label{Approach:VFN}
\vspace{-0.2cm}

The VFN approach is built upon the 2D segmentation volumes from three orthogonal ({\em coronal}, {\em sagittal} and {\em axial}) planes. Powered by state-of-the-art deep networks, these results are generally accurate ({\em e.g.}, an average DSC of over $82\%$~\cite{zhou2017fixed} on the NIH pancreas segmentation dataset~\cite{roth2015deeporgan}). But, as shown in Figure~\ref{Fig:Examples}, some {\em local} errors still occur because $2$ out of $3$ views fail to detect the target. Our assumption is that these errors can be recovered by learning and exploiting the 3D image patterns in its surrounding region.

Regarding other choices, majority voting obviously cannot take image patterns into consideration. The STAPLE algorithm~\cite{asman2013non}, while being effective in multi-atlas registration, does not have a strong ability of fitting image patterns from training data. We shall see in experiments that STAPLE is unable to improve segmentation accuracy over majority voting.

Motivated by the need to learn {\em local} patterns, we equip VFN with a small input region ($64^3$) and a shallow structure, so that each neuron has a small receptive field (the largest region seen by an output neuron is $50^3$). In comparison, in the 3D network VNet~\cite{milletari2016v}, these numbers are $128^3$ and $551^3$, respectively. This brings twofold benefits. First, we can sample more patches from the training data, and the number of parameters is much less, and so the risk of over-fitting is alleviated. Second, VFN is more computationally efficient than 3D networks, {\em e.g.}, adding 2D segmentation, it needs only half the testing time of~\cite{zhu20173d}.

\begin{figure}[!t]
\begin{center}
    \includegraphics[width=12cm]{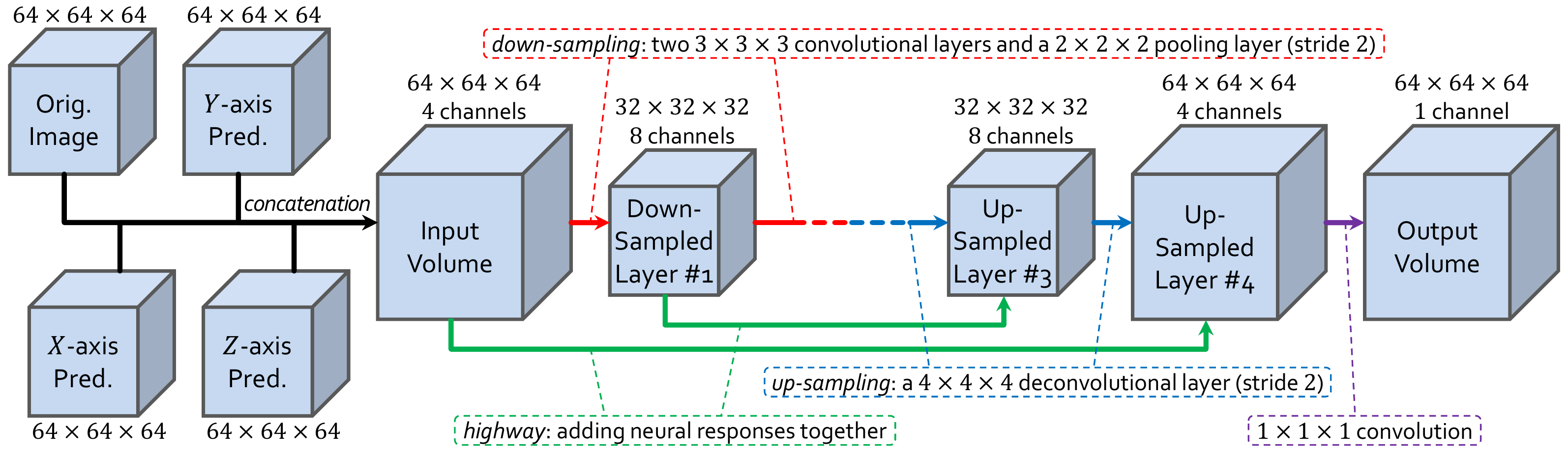}
\end{center}
\vspace{-0.5cm}
\caption{
    The network structure of VFN (best viewed in color). We only display one down-sampling and one up-sampling stages, but there are $3$ of each. Each down-sampling stage shrinks the spatial resolution by $1/2$ and doubles the number of channels. We build $3$ highway connections ($2$ are shown). We perform batch normalization and ReLU activation after each convolutional and deconvolutional layer.
}
\label{Fig:VFN}
\vspace{-0.4cm}
\end{figure}

The architecture of VFN is shown in Figure~\ref{Fig:VFN}. It has three down-sampling stages and three up-sampling stages. Each down-sampling stage is composed of two $3\times3\times3$ convolutional layers and a $2\times2\times2$ max-pooling layer with a stride of $2$, and each up-sampling stage is implemented by a single $4\times4\times4$ deconvolutional layer with a stride of $2$. Following other 3D networks~\cite{milletari2016v}\cite{zhu20173d}, we also build a few residual connections~\cite{he2016deep} between hidden layers of the same scale. For our problem, this enables the network to preserve a large fraction of 2D network predictions (which are generally of good quality) and focus on refining them (note that if all weights in convolution are identical, then VFN is approximately equivalent to majority voting). Experiments show that these highway connections lead to faster convergence and higher accuracy. A final convolution of a $1\times1\times1$ kernel reduces the number of channels to $1$.

The input layer of VFN consists of $4$ channels, $1$ for the original image and $3$ for 2D segmentations from different viewpoints. The input values in each channel are normalized into $\left[0,1\right]$. By this we provide equally-weighted information from the original image and 2D multi-view segmentation results, so that VFN can fuse them at an early stage and learn from data automatically. We verify in experiments that image information is important -- training a VFN without this input channel shrinks the average accuracy gain by half.

\vspace{-0.4cm}
\subsection{Training and Testing VFN}
\label{Approach:TrainingTesting}
\vspace{-0.2cm}

We train VFN from scratch, {\em i.e.}, all weights in convolution are initialized as random white noises. Note that setting all weights as $1$ mimics majority voting, and we find that both ways of initialization lead to similar testing performance. All $64\times64\times64$ volumes are sampled from the region-of-interest (ROI) of each training case, defined as the bounding box covering all foreground voxels padded by $32$ pixels in each dimension. We introduce data augmentation by performing random $90^\circ$-rotation and flip in 3D space (each cube has $24$ variants). We use a Dice loss to avoid background bias (a voxel is more likely to be predicted as background, due to the majority of background voxels in training). We train VFN for $30\rm{,}000$ iterations with a mini-batch size of $16$. We start with a learning rate of $0.01$, and divide it by $10$ after $20\rm{,}000$ and $25\rm{,}000$ iterations, respectively. The entire training process requires approximately $6$ hours in a Titan-X-Pascal GPU. In the testing process, we use a sliding window with a stride of $32$ in the ROI region (the minimal 3D box covering all foreground voxels of multi-plane 2D segmentation fused by majority voting). For an average pancreas in the NIH dataset~\cite{roth2015deeporgan}, testing VFN takes around $5$ seconds.

An important issue in optimizing VFN is to construct the training data. Note that we cannot reuse the data used for training segmentation networks to train VFN, because this will result in the input channels contain very accurate segmentation, which limits VFN from learning meaningful local patterns and generalizing to the testing scenarios. So, we further split the training set into two subsets, one for training the 2D segmentation networks and the other for training VFN with the testing segmentation results.

However, under most circumstances, the amount of training data is limited. For example, in the NIH pancreas segmentation dataset, each fold in cross-validation has only $60$ training cases. Partitioning it into two subsets harms the accuracy of both 2D segmentation and fusion. To avoid this, we suggest a {\bf cross-cross-augmentation} (CCA) strategy, described as follows. Suppose we split data into $K$ folds for cross-validation, and the $k_1$-th fold is left for testing. For all ${k_2}\neq{k_1}$, we train 2D segmentation models on the folds in $\left\{1,2,\ldots,K\right\}\setminus\left\{k_1,k_2\right\}$, and test on the $k_2$-th fold to generate training data for the VFN. In this way, all data are used for training both the segmentation model and the VFN. The price is that a total of $K\left(K-1\right)/2$ extra segmentation models need to be trained, which is more costly than training $K$ models in a standard cross-validation. In practice, this strategy improves the average segmentation accuracy by $\sim1\%$ in each fold. Note that we perform CCA only on the NIH dataset due to the limited amount of data -- in our own dataset, we perform standard training/testing split, requiring $<10\%$ extra training time and ignorable extra testing time.

\vspace{-0.4cm}
\section{Experiments}
\label{Experiments}

\vspace{-0.4cm}
\subsection{The NIH Pancreas Segmentation Dataset}
\label{Experiments:NIH}
\vspace{-0.2cm}

We first evaluate our approach on the NIH pancreas segmentation dataset~\cite{roth2015deeporgan} containing $82$ abdominal CT volumes. The width and height of each volume are both $512$, and the number of slices along the {\em axial} axis varies from $181$ to $466$. We split the dataset into $4$ folds of approximately the same size, and apply cross-cross-augmentation (see Section~\ref{Approach:TrainingTesting}) to improve segmentation accuracy.

\newcommand{\colwidthA}{2.0cm}
\newcommand{\colwidthB}{0.9cm}
\newcommand{\colwidthC}{1.4cm}
\begin{table}[!btp]
\centering
\begin{tabular}{|l||R{\colwidthA}|R{\colwidthB}|R{\colwidthB}|R{\colwidthB}|R{\colwidthB}|R{\colwidthB}||R{\colwidthC}|}
\hline
Approach                                   &                  Average &     Min & $1/4$-Q &     Med & $3/4$-Q &     Max & Time (m) \\
\hline\hline
Roth {\em et al.}~\cite{roth2015deeporgan} &          $71.42\pm10.11$ & $23.99$ &     $-$ &     $-$ &     $-$ & $86.29$ & $6$--$8$ \\
\hline
Roth {\em et al.}~\cite{roth2016spatial}   &          $78.01\pm 8.20$ & $34.11$ &     $-$ &     $-$ &     $-$ & $88.65$ & $2$--$3$ \\
\hline
Roth {\em et al.}~\cite{roth2017spatial}   &          $81.27\pm 6.27$ & $50.69$ &     $-$ &     $-$ &     $-$ & $88.96$ & $2$--$3$ \\
\hline
Cai {\em et al.}~\cite{cai2017pancreas}    &          $82.4 \pm 6.7 $ & $60.0 $ &     $-$ &     $-$ &     $-$ & $90.1 $ &     N/A \\
\hline
Zhu {\em et al.}~\cite{zhu20173d}          &          $84.59\pm 4.86$ & $69.62$ &     $-$ &     $-$ &     $-$ & $91.45$ &    $4.1$ \\
\hline\hline
Zhou {\em et al.}~\cite{zhou2017fixed}     &          $82.50\pm 6.14$ & $56.33$ & $81.63$ & $84.11$ & $86.28$ & $89.98$ &    $0.9$ \\
\hline
\cite{zhou2017fixed} + NLS                 &          $82.25\pm 6.57$ & $56.86$ & $81.54$ & $83.96$ & $86.14$ & $89.94$ &    $1.1$ \\
\hline
\cite{zhou2017fixed} + VFN                 & $\mathbf{84.06}\pm 5.63$ & $62.93$ & $81.98$ & $85.69$ & $87.62$ & $91.28$ &    $1.0$ \\
\hline\hline
Yu {\em et al.}~\cite{yu2017recurrent}     &          $84.48\pm 5.03$ & $62.23$ & $82.50$ & $85.66$ & $87.82$ & $91.17$ &    $1.3$ \\
\hline
\cite{yu2017recurrent} + NLS               &          $84.47\pm 5.03$ & $62.22$ & $82.42$ & $85.59$ & $87.78$ & $91.17$ &    $1.5$ \\
\hline
\cite{yu2017recurrent} + VFN               & $\mathbf{84.63}\pm 5.07$ & $61.58$ & $82.42$ & $85.84$ & $88.37$ & $91.57$ &    $1.4$ \\
\hline
\end{tabular}
\caption{
    Comparison of segmentation accuracy (DSC, $\%$) and testing time (in minutes) between our approach and the state-of-the-arts on the NIH dataset~\cite{roth2015deeporgan}. \cite{zhou2017fixed} and~\cite{yu2017recurrent} are reimplemented by ourselves, and the default fusion is majority voting.
}
\label{Tab:ComparisonNIH}
\vspace{-0.8cm}
\end{table}

\begin{figure}[t]
\begin{center}
    \includegraphics[width=12cm]{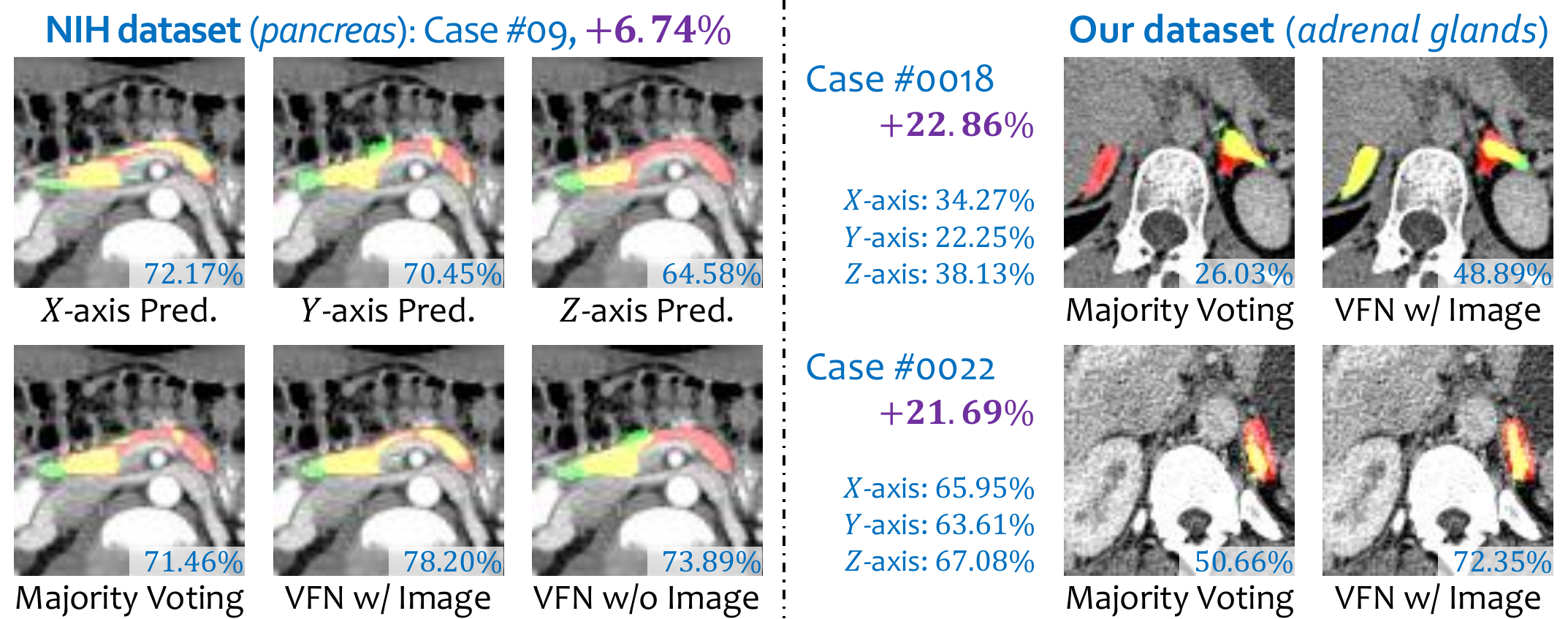}
\end{center}
\vspace{-0.5cm}
\caption{
    Two typical examples, each with the original image, segmentation results from three viewpoints, and different fusion results. In each label map, red, green and yellow indicate ground-truth, prediction and overlap, respectively (best viewed in color).
}
\label{Fig:Examples}
\vspace{-0.4cm}
\end{figure}

Results are summarized in Table~\ref{Tab:ComparisonNIH}. We use two recent 2D segmentation approaches as our baseline, and compare VFN with two other fusion approaches, namely majority voting and non-local STAPLE (NLS)~\cite{asman2013non}. The latter was verified more effective than its former local version. We measure segmentation accuracy using DSC and report the average accuracy over $82$ cases. Based on~\cite{zhou2017fixed}, VFN improves majority voting significantly by an average of $1.69\%$. The improvement over $82$ cases is consistent (the student's $t$-test reports a $p$-value of $6.9\times10^{-7}$), although the standard deviation over $82$ cases is relatively large -- this is mainly caused by the difference in difficulties from case to case. Figure~\ref{Fig:Examples} shows an example on which VFN produces a significant accuracy gain. VFN does not improve~\cite{yu2017recurrent} significantly, arguably because~\cite{yu2017recurrent} has almost reached the human-level agreement (we invited a radiologist to segment this dataset individually, and she achieves an average accuracy of $\sim86\%$). Note that the other approaches without CCA used both the training and validation folds for training, and so all numbers are comparable in Table~\ref{Tab:ComparisonNIH}.

Due to our analysis in Section~\ref{Approach:VFN}, NLS does not produce any accuracy gain over either~\cite{zhou2017fixed} and~\cite{yu2017recurrent}. NLS is effective in multi-atlas registration, where the labels come from different images and the annotation is relatively accurate~\cite{asman2013non}. But in our problem, segmentation results from 2D networks can be noisy, thus recovering these errors requires learning local image patterns from training data, which is what VFN does to outperform NLS.

To reveal the importance of image information, we train a VFN without the image channel in the input layer. Based on~\cite{zhou2017fixed}, this version produces approximately half of the improvement ($1.69\%$) by the full model. We show an example in Figure~\ref{Fig:Examples}, in which the right part of the pancreas is missing in both {\em sagittal} and {\em axial} planes, but the high confidence in the {\em coronal} plane and the continuity of image intensities suggest its presence in the final segmentation.

\vspace{-0.4cm}
\subsection{Our Multi-Organ Dataset}
\label{Experiments:Ours}
\vspace{-0.2cm}

The radiologists in our team collected a dataset with $300$ high-resolution CT scans. These scans were performed on some potential renal donors. Four experts in abdominal anatomy annotated $11$ abdominal organs, taking $3$--$4$ hours for each scan, and all annotations were verified by an experienced board certified Abdominal Radiologist. Except for the {\em pancreas}, we choose several challenging targets, including the {\em adrenal glands}, the {\em duodenum}, and the {\em gallbladder} (easy cases such as the {\em liver} and the {\em kidneys} are not considered). We use $150$ cases for training 2D segmentation models, $100$ cases for training VFN, and test on the remaining $50$ cases. The data split is random but identical for different organs.

\renewcommand{\colwidthA}{2.0cm}
\begin{table}[!btp]
\centering
\begin{tabular}{|l||R{\colwidthA}|R{\colwidthA}|R{\colwidthA}|R{\colwidthA}|R{\colwidthA}|}
\hline
Approach                               &         {\em adrenal g.} &           {\em duodenum} &        {\em gallbladder} &           {\em pancreas} \\
\hline\hline
Zhu {\em et al.}~\cite{zhu20173d}      &          $36.74\pm25.14$ &          $68.80\pm14.38$ &          $42.01\pm29.47$ &          $85.25\pm 6.04$ \\
\hline\hline
Zhou {\em et al.}~\cite{zhou2017fixed} &          $66.09\pm18.19$ &          $71.65\pm13.15$ &          $90.39\pm 5.30$ &          $84.52\pm 6.23$ \\
\hline
\cite{zhou2017fixed} + VFN             & $\mathbf{69.24}\pm17.42$ & $\mathbf{72.77}\pm12.80$ & $\mathbf{91.40}\pm 5.19$ & $\mathbf{86.39}\pm 6.20$ \\
\hline\hline
Yu {\em et al.}~\cite{yu2017recurrent} &          $71.40\pm12.87$ &          $77.48\pm 8.70$ &          $91.81\pm 4.90$ &          $87.22\pm 5.90$ \\
\hline
\cite{yu2017recurrent} + VFN           & $\mathbf{72.09}\pm13.61$ & $\mathbf{77.77}\pm 8.46$ & $\mathbf{92.15}\pm 5.05$ & $\mathbf{88.06}\pm 5.33$ \\
\hline
\end{tabular}
\caption{
    Comparison of segmentation accuracy (DSC, $\%$) on our multi-organ dataset. The baseline for~\cite{zhou2017fixed} and~\cite{yu2017recurrent} is majority voting. The numbers of~\cite{yu2017recurrent} are different from those in their original paper, because we are using a different dataset.
}
\label{Tab:ComparisonOurs}
\vspace{-0.8cm}
\end{table}

Results are shown in Table~\ref{Tab:ComparisonOurs}. Again, our approach consistently improves 2D segmentation, which demonstrates the transferability of our methodology. In {\em pancreas}, based on~\cite{yu2017recurrent}, we obtain a $p$-value of $2.7\times10^{-5}$ over $50$ testing cases. In {\em adrenal glands}, although the average accuracy gains are not large, the improvement is significant in some badly segmented cases, {\em e.g.}, Figure~\ref{Fig:Examples} shows two examples with more than $20\%$ accuracy boosts. Refining bad segmentations makes our segmentation results more reliable. By contrast, the 3D network~\cite{zhu20173d} produces unstable performance ({\cite{zhu20173d} was designed for pancreas segmentation, thus works reasonably well in {\em pancreas}), which is mainly caused by the limited training data especially for small organs such as {\em adrenal glands} and {\em gallbladder}.

Therefore, we conclude that 2D segmentation followed by 3D fusion is currently a very promising idea to bridge the gap between 2D and 3D segmentation approaches, particularly if there is limited training data.

\vspace{-0.4cm}
\section{Conclusions}
\label{Conclusions}
\vspace{-0.2cm}

In this paper, we discuss an important topic in medical imaging analysis, namely bridging the gap between 2D and 3D organ segmentation approaches. We propose to train more stable 2D segmentation networks, and then use a light-weighted 3D fusion module to fuse their results. In this way, we enjoy the benefits of exploiting 3D information to improve segmentation, as well as avoiding the risk of over-fitting caused by tuning 3D models (which have $10\times$ more parameters) on a limited amount of training data. We verify the effectiveness of our approach on two datasets, one of which contains several challenging organs.

Based on our work, a promising direction is to train the segmentation and fusion modules in a joint manner, so that the 2D networks can incorporate 3D information in the training process by learning from the back-propagated gradients of VFN. Another issue involves training VFN more efficiently, {\em e.g.}, using hard example mining. These topics are left for future research.

\vspace{0.2cm}
\noindent
{\bf Acknoledgements} This work was supported by the Lustgarten foundation for pancreatic cancer research. We thank Prof. Seyoun Park, Prof. Wei Shen, Dr. Yan Wang and Yuyin Zhou for instructive discussions.
\vspace{-0.2cm}

\bibliographystyle{splncs03}
\bibliography{typeinst}

\begin{thebibliography}{10}
\providecommand{\url}[1]{\texttt{#1}}
\providecommand{\urlprefix}{URL }

\bibitem{asman2013non}
Asman, A.J., Landman, B.A.: Non-local statistical label fusion for multi-atlas
  segmentation. Medical Image Analysis  17(2),  194--208 (2013)

\bibitem{boykov2000interactive}
Boykov, Y., Jolly, M.P.: Interactive organ segmentation using graph cuts. In:
  MICCAI (2000)

\bibitem{cai2017pancreas}
Cai, J., Lu, L., Xie, Y., Xing, F., Yang, L.: Pancreas segmentation in mri
  using graph-based decision fusion on convolutional neural networks. In:
  MICCAI (2017)

\bibitem{cicek20163d}
Cicek, O., Abdulkadir, A., Lienkamp, S.S., Brox, T., Ronneberger, O.: 3d u-net:
  learning dense volumetric segmentation from sparse annotation. In: MICCAI
  (2016)

\bibitem{he2016deep}
He, K., Zhang, X., Ren, S., Sun, J.: Deep residual learning for image
  recognition. In: CVPR (2016)

\bibitem{krizhevsky2012imagenet}
Krizhevsky, A., Sutskever, I., Hinton, G.E.: Imagenet classification with deep
  convolutional neural networks. In: NIPS (2012)

\bibitem{long2015fully}
Long, J., Shelhamer, E., Darrell, T.: Fully convolutional networks for semantic
  segmentation. In: CVPR (2015)

\bibitem{milletari2016v}
Milletari, F., Navab, N., Ahmadi, S.A.: V-net: Fully convolutional neural
  networks for volumetric medical image segmentation. In: 3DV (2016)

\bibitem{roth2015deeporgan}
Roth, H.R., Lu, L., Farag, A., Shin, H., Liu, J., Turkbey, E.B., Summers, R.M.:
  Deeporgan: Multi-level deep convolutional networks for automated pancreas
  segmentation. In: MICCAI (2015)

\bibitem{roth2016spatial}
Roth, H.R., Lu, L., Farag, A., Sohn, A., Summers, R.M.: Spatial aggregation of
  holistically-nested networks for automated pancreas segmentation. In: MICCAI
  (2016)

\bibitem{roth2017spatial}
Roth, H.R., Lu, L., Lay, N., Harrison, A.P., Farag, A., Sohn, A., Summers,
  R.M.: Spatial aggregation of holistically-nested convolutional neural
  networks for automated pancreas localization and segmentation.
  arXiv:1702.00045  (2017)

\bibitem{setio2016pulmonary}
Setio, A.A.A., Ciompi, F., Litjens, G., Gerke, P., Jacobs, C., van Riel, S.J.,
  Wille, M.M.W., Naqibullah, M., S{\'a}nchez, C.I., van Ginneken, B.: Pulmonary
  nodule detection in ct images: false positive reduction using multi-view
  convolutional networks. IEEE TMI  35(5),  1160--1169 (2016)

\bibitem{su2015multi}
Su, H., Maji, S., Kalogerakis, E., Learned-Miller, E.: Multi-view convolutional
  neural networks for 3d shape recognition. In: ICCV (2015)

\bibitem{tajbakhsh2016convolutional}
Tajbakhsh, N., Shin, J.Y., Gurudu, S.R., Hurst, R.T., Kendall, C.B., Gotway,
  M.B., Liang, J.: Convolutional neural networks for medical image analysis:
  Full training or fine tuning? IEEE TMI  35(5),  1299--1312 (2016)

\bibitem{tu2010auto}
Tu, Z., Bai, X.: Auto-context and its application to high-level vision tasks
  and 3d brain image segmentation. IEEE TPAMI  32(10),  1744--1757 (2010)

\bibitem{yang2016deep}
Yang, H., Sun, J., Li, H., Wang, L., Xu, Z.: Deep fusion net for multi-atlas
  segmentation: Application to cardiac mr images. In: MICCAI (2016)

\bibitem{yu2017recurrent}
Yu, Q., Xie, L., Wang, Y., Zhou, Y., Fishman, E.K., Yuille, A.L.: Recurrent
  saliency transformation network: Incorporating multi-stage visual cues for
  small organ segmentation. arXiv:1709.04518  (2017)

\bibitem{zhou2017fixed}
Zhou, Y., Xie, L., Shen, W., Wang, Y., Fishman, E.K., Yuille, A.L.: A
  fixed-point model for pancreas segmentation in abdominal ct scans. In: MICCAI
  (2017)

\bibitem{zhu20173d}
Zhu, Z., Xia, Y., Shen, W., Fishman, E.K., Yuille, A.L.: A 3d coarse-to-fine
  framework for automatic pancreas segmentation. arXiv:1712.00201  (2017)

\end{thebibliography}
\end{document}